\documentclass[a4paper]{article}
\pdfoutput=1

\usepackage{graphicx}
\usepackage{bm}								
\usepackage{algorithm}
\usepackage{algpseudocode}
\usepackage[english]{babel}   
\usepackage{enumerate} 				
\usepackage{paralist}					
\usepackage{amsfonts}
\usepackage{amsmath}   				
\usepackage{tikz}	
\usetikzlibrary{arrows,positioning,shapes,matrix,calc} 
\usepackage{natbib}
\setcitestyle{authoryear,open={[},close={]}}

\newcommand{\latexOrPdflatex}[2]{\ifx\undefined\pdfoutput%
#1%
\else%
#2%
\fi}

\latexOrPdflatex{
\newcommand{\href}[2]{#2}
}{
\usepackage[colorlinks,pagebackref,citecolor=blue,bookmarks=true,pdfstartview=FitH,pdfstartpage=1,pdftitle={Driving Forces and SFA},pdfauthor={Wolfgang Konen}]{hyperref}
}

\title{%
\vspace{-0.5cm}{\small \tt \raggedleft{e-print published at
\href{http://arxiv.org}{http://arxiv.org}, March 2016}} \\
\vspace{1.cm} 
Illumination-invariant image mosaic calculation based on logarithmic search
}
\author{Wolfgang Konen \\ \\
 Institute for Computer Science \\
 Cologne University of Applied Sciences \\
 Steinm\"ullerallee 1, D-51643 Gummersbach, Germany \\
 March 2016\\
 \texttt{\href{mailto:{wolfgang.konen,patrick.koch}@fh-koeln.de}{wolfgang.konen@fh-koeln.de}}
}

\date{}

\begin{document} 

\maketitle 

\begin{abstract}
This technical report describes an improved image mosaicking algorithm. It is based on Jain's logarithmic search algorithm~\citep{jain1981displacement} which is coupled to the method of \citet{kourogi1999real} for matching images in a video sequence. Logarithmic search has a better invariance against illumination changes than the original optical-flow-based method of Kourogi.

\end{abstract}



\section{Introduction} \label{sec:introduction}

\subsection{Related work}
Many algorithms on image mosaicking are known, however only relatively few of them can work fully automatic, e.g. \citep{kourogi1999real,szeliski1994image,seshamani2006real,robinson2003simplex} and under real-time conditions \citep{kourogi1999real,seshamani2006real,robinson2003simplex}.  Except for \cite{seshamani2006real} they have not yet been applied to endoscopic video. 

A problem with Kourogi's method \citep{kourogi1999real,KonBre07-Real,KonNad07} is its dependence on illumination conditions. It is a common condition of endoscopic video sequences that the light source moves together with the camera so that the illumination of a certain patch will vary with its location in the field of view. The logarithmic search method~\citep{jain1981displacement,Lundmark2001LogSearch} is a fast search method based on normalized cross-correlation and thus very robust against overall intensity differences or against slowly varying gradients. This report describes a synthesis between logarithmic search and Kourogi's algorithm.

\section{Methods}
\subsection{Normalized cross correlation}
\label{sec:CrossCorrel}
The normalized cross correlation is a robust measure to quantify the similarity between two image patches (or templates) $A$ and $B$. In the most simple form, these patches can be rectangular areas within images, but irregular shapes are possible as well. The normalized cross correlation is defined as
\begin{equation}
		C_L(A,B) = \frac{\sum_{m,n} (A_{mn}-\bar{A})(B_{mn}-\bar{B})}
							 {\sqrt{(\sum_{m,n} (A_{mn}-\bar{A})^2)(\sum_{m,n} (B_{mn}-\bar{B})^2)}}
\label{eq:crosscorr}
\end{equation}
where $\bar{A}$ and $\bar{B}$ are the average intensities of patch $A$ and $B$, resp. Due to average subtraction and normalization to the variance of both patches, the measure $C_L$ is invariant against global intensity or contrast changes and only weakly affected by a slowly varying intensity gradient. The correlation coefficient $C_L$ is in the range $[-1,1]$. 

\subsection{Logarithmic search}
\label{sec:LogSearch}

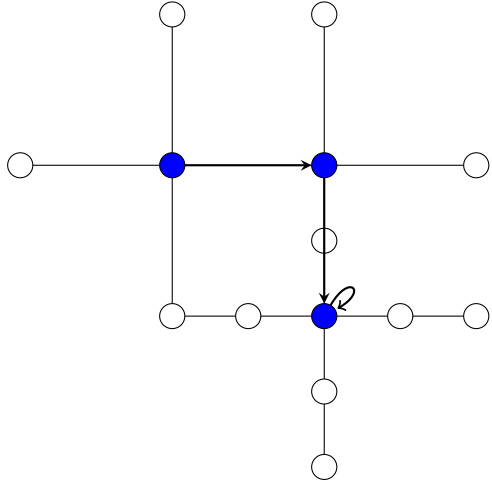
\begin{figure}%
\begin{center}
\begin{tikzpicture}
	%
	%
	\matrix (network)
	[matrix of nodes,%
	nodes in empty cells,
	nodes={outer sep=0pt,circle,minimum size=4pt,draw},
	column sep={1cm,between origins},
	row sep={1cm,between origins}]
	{
		|[draw=none]| & |[draw=none]| &               & |[draw=none]| &               & |[draw=none]| & |[draw=none]| \\
		|[draw=none]| & |[draw=none]| & |[draw=none]| & |[draw=none]| & |[draw=none]| & |[draw=none]| & |[draw=none]| \\
		              & |[draw=none]| & |[fill=blue]| & |[draw=none]| & |[fill=blue]| & |[draw=none]| &               \\
		|[draw=none]| & |[draw=none]| & |[draw=none]| & |[draw=none]| &               & |[draw=none]| & |[draw=none]| \\
		|[draw=none]| & |[draw=none]| &               &               & |[fill=blue]|  & &               \\
		|[draw=none]| & |[draw=none]| & |[draw=none]| & |[draw=none]| &               & |[draw=none]| & |[draw=none]| \\
		|[draw=none]| & |[draw=none]| & |[draw=none]| & |[draw=none]| &               & |[draw=none]| & |[draw=none]| \\
	};
	%
	%
	\draw (network-3-1) -- (network-3-3);
	\draw (network-3-3) -- (network-3-5);
	\draw (network-3-5) -- (network-3-7);
	
	\foreach \b in {3,5}{
			\draw (network-1-\b) -- (network-3-\b);
			\draw (network-3-\b) -- (network-5-\b);
	}
	\draw (network-5-3) -- (network-5-4);
	\draw (network-5-4) -- (network-5-5);
	\draw (network-5-5) -- (network-5-6);
	\draw (network-5-6) -- (network-5-7);
	\draw (network-5-5) -- (network-6-5);
	\draw (network-6-5) -- (network-7-5);
	\draw [-stealth,thick] (network-3-3) -- (network-3-5);
	\draw [-stealth,thick] (network-3-5) -- (network-5-5);
	\draw [-stealth,thick] (network-5-5) edge [in=30,out=60,loop]  (network-5-5);
\end{tikzpicture}	
\end{center}
\caption{Example of a logarithmic search}%
\label{fig:LogSearch}%
\end{figure}

Logarithmic search~\citep{jain1981displacement} is an improvement over naive template search. If the search area is a square of side length $2w$, then a naive template search requires $4w^2$ cross correlation measurements. The logarithmic search instead forms a cross with cross bar length $w$ and either shifts this cross or halves the cross bar until it reaches $w=1$. Fig.~\ref{fig:LogSearch} shows an example where the cross is first shifted to the right, then down, where it finds the best correlation in the center of the cross. Therefore the cross bar length $w$ is halved, and the search continues with the smaller cross. It is easy to see that the complexity of this approach is only  $5 (ld(w)+S)$ where $S$ is the number of times we have to shift the cross, which is usually a small number.\footnote{Instead of a 5-point cross it is equally well possible to use a 9-point neighborhood (including self). 
} The complexity reduction from $O(w^2)$ to $O(ld(w))$ is the reason for the name \glqq logarithmic search\grqq.

Algorithm~\ref{alg:logSearch} describes logarithmic search in detail. Usually template $A$ will be a square, but other shapes are possible as well. Template $B$ has to have the same shape and size as $A$. \glqq Extract template $A$ at location $(x,y)$\grqq\ in line 2 means that the center location of the square is at $(x,y)$. Similar for template $B$ in line 6. The initial size $w_{init}$ is usually a power of 2, so that every division by 2 in line 9 yields again an integer value. 

If the cross correlation function $C_L(A,B)$ is convex in the search area, logarithmic search is guaranteed to converge to the correct (and more costly to compute) solution of the naive template search.

\begin{algorithm}[tbp]%
\caption{Logarithmic search. \textbf{Input:} Reference image $I_r$, object image $I_o$, 
landmark $L=(x,y)$ in reference image, compensated motion $(u_c,v_c)$, initial legnth $w_{init}$ of cross bar. 
\textbf{Output:} The best location $(u,v)$ in object image and the corresponding normalized cross correlation $C_L(A,B)$.}
\label{alg:logSearch}
	\begin{algorithmic}[1] 
\Function{LogSearch}{$L,u_c,v_c,I_r,I_o$}
\State Extract template $A$ at location $L = (x,y)$ from reference image $I_r$.
\State $(u,v) \leftarrow (x+u_c,y+v_c)$    \Comment{initial position }
\State $w \leftarrow w_{init}$				\Comment{$w_{init}$ is the initial length of each cross bar} 
\While{$w>1$}
	\State Extract template $B$ at the five locations $(u,v),(u+w,v),(u,v+w),(u-w,v),(u,v-w)$  from image $I_o$ and measure $C_L(A,B)$ for each location      \Comment{the locations form a cross around $(u,v)$}
	\State Set $(u_{max},v_{max}) \leftarrow$ location where $C_L(A,B)$ is maximal
	\If{$(u_{max},v_{max}) == (u,v)$}
			\State $w \leftarrow w/2$
	\Else															\Comment{shift the cross to $(u_{max},v_{max})$}
			\State $(u,v) \leftarrow (u_{max},v_{max})$
	\EndIf
\EndWhile
\State \Return $(u,v,C_L(A,B(u,v)))$
\EndFunction

\end{algorithmic}
\end{algorithm}%

\subsection{Kourogi's pseudo motion method}
\label{sec:kourogi}
The pseudo motion algorithm according to \citet{kourogi1999real} -- based on the seminal work of \citet{horn1981determining} on optical flow -- is described in more detail in~\citet{Kon09-TRKourogi} (in German) and in \citet{KonBre07-Real,KonNad07}. It estimates a compensated motion vector field $(u_c,v_c)$ for each pixel in the image. Usually it starts with a compensated motion equal to $(0,0)$ at each pixel and fills in a dense compensated motion field through a loop of iterations.

For reference, we describe this algorithm in a nutshell as follows:
The goal of the improved optical flow algorithm according to \cite{kourogi1999real} is to estimate the motion field between successive frames $I(t-1)$ and $I(t)$ of a video sequence. It calculates at each pixel $(x,y)$ the so-called pseudo motion 
\begin{equation}
		\left(
		\begin{matrix} u_p \\ v_p \end{matrix} 
		\right)  =
		\left(
		\begin{matrix} - I_t^{(c)}/I_x \\ - I_t^{(c)}/I_y \end{matrix} 
		\right)  +
		\left(
		\begin{matrix} u_c \\ v_c \end{matrix} 
		\right)  
		\quad\mbox{with}\quad
		I_t^{(c)} = I(x+u_c,y+v_c,t) - I(x,y,t-1)
\label{eq:pseudoMotion}
\end{equation}
where $I_x$ and $I_y$ denote the spatial gradient and $(u_c,v_c)$ is the so-called compensated motion at this pixel location. $I(x,y,t)$ is the luminance signal at pixel $(x,y)$ in frame $t$. Our algorithm proceeds as follows: Initially we start with $(u_c,v_c)=0$ or with an estimate from the previous frame. Then the following steps are carried out in a loop:
\begin{enumerate}[(A)]				
	\item Calculate the pseudo motion $(u_p,v_p)$ according to Eq.~\eqref{eq:pseudoMotion} for each pixel inside the endoscopic mask.
	\item Accept only those pixel which fulfill the following criteria: (a) $I_x$ and $I_y$ are not 0, (b) $(x+u_p,y+v_p)$ is inside the endoscopic mask and (c) $|I(x+u_p,y+v_p,t) - I(x,y,t-1)| < T$. Here, $T$ is a suitable gray level threshold, e.~g. $T=5$.
	\item Find the affine parameters $\vec{a} = \{a_1,\ldots,a_6\}$ for a global motion field best-fitting the pseudo motion at all accepted pixel locations $i$, i.~e. solve the overdetermined system of equations
\begin{equation}
		\begin{matrix} 
				a_1 x_i + a_2 y_i + a_3 = u_{p,i} \\ 
				a_4 x_i + a_5 y_i + a_6 = v_{p,i}
		\end{matrix} 
\label{eq:LSequations}
\end{equation}
in a least-squares sense.\footnote{It is also possible to use instead of the 6-parameter affine transformation the 8-parameter projective transformation, if the image material requires this wider class of transformations.} The solution $\vec{a}$ from Eq.~\eqref{eq:LSequations} gives a new estimate for $(u_c,v_c)$ 
\begin{equation}
		\begin{matrix} 
				u_c(x_i,y_i) = a_1 x_i + a_2 y_i + a_3  \\ 
				v_c(x_i,y_i) = a_4 x_i + a_5 y_i + a_6 
		\end{matrix} 
\label{eq:uc_vc}
\end{equation}
\item Continue with step (A) using the new compensated motion vector field $(u_c,v_c)$.
\end{enumerate}
The loop is terminated either after a fixed number of iterations or when the change in the global motion field drops below a certain threshold.

In contrast to that, the pseudo motion method based on logarithmic search, which will be described next, is a single-pass algorithm (no iterations). It calculates the precise pseudo motion vectors only for a set of landmarks, not for all pixel. 

\subsection{Pseudo motion based on logarithmic search}
\label{sec:PMotion}

\begin{algorithm}%
\caption{Pseudo motion with logarithmic search. \textbf{Input:} Reference image $I_r$, object image $I_o$, threshold $c_{min}$ for cross correlation coefficient, minimum acceptance rate $a_{min}$, maximum distance $e_{max}$. 
\textbf{Output:} The best transformation $\vec{a}$ and the set of accecpted matches $M''$.}
\label{alg:pmotion_log}
	\begin{algorithmic}[1] 
\Function{PMotionLog}{$I_r,I_o,c_{min},a_{min},e_{max},N$}
\State Initial estimate of compensated motion $(u_c,v_c)$
\State Distribute $N$ landmarks $L$ over the reference image
\State $M = \{\}$ 										\Comment{LogSearch part}
\For {(each landmark $L = (x,y)$)}
	\State Find corresponding location in object image: $(u_p,v_p,C_L) = $\Call{LogSearch}{$L,u_c,v_c,I_r,I_o$} 
	\State $M \leftarrow M \cup \{(L,u_p,v_p,C_L)\}$
\EndFor
\State Sort the tuples in $M$ by $C_L$ in decreasing order 										\Comment{Remove bad matches}
\State $M' = \{m \in M~|~m~\mbox{is among the first} ~a_{min}N~ \mbox{elements of the sorted list OR} ~C_L \geq c_{min} \}$ 
\State Calculate the transformation $\vec{a}$ as least-squares estimate from Eq.~\eqref{eq:LSequations} based on all tuples in $M'$. 
\State Recalculate $(u_c,v_c)$ according to Eq.~\eqref{eq:uc_vc}
\State $M = \{\}$ 										\Comment{Remove outliers}
\For {(each tuple $m \in M'$)}
	\State Calculate the Euclidean distance $e_m = \sqrt{(u_p-u_c)^2+(v_p-v_c)^2}$
	\State $M \leftarrow M \cup \{(m,e_m)\}$
\EndFor
\State Sort the tuples in $M$ by $e_m$ in increasing order 
\State $M'' = \{m \in M~|~m~\mbox{is among the first} ~a_{min}N~ \mbox{elements of the sorted list OR} ~e_m < e_{max} \}$ 
\State Calculate the transformation $\vec{a}$ as least-squares estimate from Eq.~\eqref{eq:LSequations} based on all tuples in $M''$. 
\State \Return $(\vec{a}, M'')$
\EndFunction
\end{algorithmic}
\end{algorithm}%

Algorithm~\ref{alg:pmotion_log} describes the new pseudo motion method in detail. Step 2 is an initial estimate of the compensated motion. Several possibilities for such an estimate exist: 
\begin{compactenum}
	\item Initialize $(u_c,v_c)=0$ for every pixel. 
	\item Initialize $(u_c,v_c)$ with the compensated motion from the previous frame according to Eq.~\eqref{eq:uc_vc}.
	\item Call Kourogi's method for a small number of iterations to get an initial estimate of $(u_c,v_c)$.\footnote{To avoid any instabilities in this initialization method, we usually restrict Kourogi's method to the class of purely translational transformations.}
\end{compactenum}

In any way, we start with an initial estimate for $(u_c,v_c)$ and define a set of $N$ landmarks. We use in line 6 the 
logarithmic search method to relocate the landmarks in the object image, starting from the displacement given by $(u_c,v_c)$. 

In order to select only reliable landmarks, there is a two-stage filter in lines 9-19 of Algorithm~\ref{alg:pmotion_log}: 
\begin{enumerate}[(1)]
	\item First we eliminate all landmarks with final correlation coefficient below a threshold, $C_L<c_{min}$. We have however a request for a minimum of $a_{min} N$ landmarks in total. If there are not enough landmarks above threshold, we take the $a_{min}N$ landmarks with highest $C_L$.
	\item Based on the landmarks accepted in the first stage, we recalculate $(u_c,v_c)$ with the help of Eq.~\eqref{eq:LSequations} and Eq.~\eqref{eq:uc_vc}. We now measure for each landmark the Euclidean distance between its displacement vector $(u_p,v_p)$ and its compensated motion $(u_c,v_c)$. Outliers (landmarks having a distance larger than $e_{max}$) are discarded. Again, care is taken to retain at least $a_{min} N$ landmarks.
\end{enumerate}

The finally accepted landmarks in $M''$ are the basis for the least-squares estimate of transformation $\vec{a}$.

Given the resulting transformation $\vec{a}$ from algorithm \textsc{PMotionLog}, the object image is now warped into the reference image frame. If there is a whole video sequence, these steps are repeated in a loop and step-by-step an image mosaic is formed. 

The actual implementation is a little bit more complicated, since only a certain masked area of the image frame contains pixel from the endoscopic view. Furthermore, valid locations $(u,v)$ for template $A$ in the reference or template $B$ in the object image are only those locations where these templates are fully inside the masked area. Those valid locations are computed beforehand with the help of suitable erosion operators.    

\begin{figure}[tbp]%
\centering
\includegraphics[width=0.3\columnwidth]{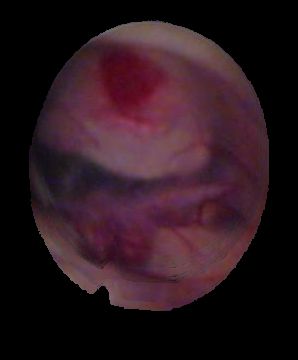} %
\includegraphics[width=0.3\columnwidth]{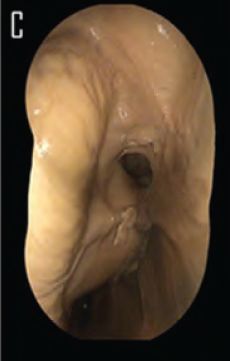} 
\caption{Examples of image mosaicking results \citep{KonNad07,Lieb14}.}%
\label{fig:mosaicResults}%
\end{figure}

\section{Results}

Fig.~\ref{fig:mosaicResults} shows some image mosaicking results. The left image is an early result with Kourogi's method from \citet{KonNad07} (CARS conference), based on an neuroendoscopic videos from an older Wolf endoscope. The right image is from the more recent study of \citet{Lieb14} where a Storz endoscope with better lighting and better image quality was available. Here the logarithmic search algorithm is used. A clear improvement in quality is visible. For details the reader is referred to the relevant publications \citep{KonNad07,Lieb14}.

\section{Discussion}

\textbf{Advantages} of pseudo motion based on logarithmic search:
\begin{itemize}
	\item It is invariant against local intensity and local contrast changes in the images.
	\item A good match is well recognizable by its high correlation coefficient $C_L$.
	\item Localization is usually more precise than with the pixel-based method of Kourogi. (Problems with the pixel-based method of Kourogi are: (a) the gray level of a pixel can fit at multiple places; (b) it may not fit at the correct place because the overall intensity in both images is different.)
	\item It avoids the problematic acceptance test in step (B) of Kourogi's method (see Sec.~\ref{sec:kourogi}).
	\item It is sufficient to match a small set of landmarks. (In principle, 3 landmarks would be sufficient to estimate the 6 parameters of $\vec{a}$. For more robust operation in practical applications, we usually take 10-20 landmarks, which is still considerable less than the number of pixels.) A small set of landmarks allows fast calculation although the individual logarithmic search method is more costly than the pixel acceptance test.
	\item Most importantly, there is no loop with 10-30 iterations as in Kourogi's method. It was found that for difficult image pairs (large distortions or difficult lighting conditions), these iterations may cause divergent solutions (e.~g. with unrealistically large scale factors). In contrast to that, logarithmic search is considerably more stable.
\end{itemize}

\noindent \textbf{Disadvantages} of pseudo motion based on logarithmic search:
\begin{itemize}
	\item It is not applicable to (larger) rotations. But in video images there are usually no large rotations from frame to frame.
	\item Cross correlation accuracy decreases if there is a larger scale factor between two images to be matched. But scale factors up to 10-15\% will be usually tolerated.
	\item It does not calculate a \textit{dense} pseudo motion field $(u_p,v_p)$ as Kourogi's method does. Instead, $(u_p,v_p)$ is only calculated for the accepted landmarks. Thus it cannot express the discontinuities in motion fields as they appear in images from natural scenes when a nearer object with sharp boundaries moves in front of a more distant background. But it was found that endoscopic video sequences often do not exhibit such discontinuities, because the near-far-transition is often gradually.  This is of course only a first approximation, and there might be other endoscopic applications, where a 3D reconstruction of nearby objects is necessary to create visually acceptable image mosaics. But for a first approach where the image mosaic gives the surgeon some orientation in the scene, it might be more important to have an algorithm which works fast and robustly. This holds for the logarithmic search method.  
\end{itemize}

Logarithmic search was also used for tracking landmarks in neuroendoscopic video sequences \citep{KonST+97bv_ess,SchKT98e,KonST98a,SchTK05}. Logarithmic search was found to work robustly and a tracking error of 0.7 mm could be obtained \citep{KonST+97bv_ess}. Tracking operates at 8 frames per second and thus can be performed under real-time conditions in the operating theatre.

Besides its applications in neuroendoscopy, logarithmic search was also applied with very good success to image stabilization of facial video sequences in the diploma thesis of \citet{Bloemen09} (in German). It was shown that it operates reliable enough to allow super-resolution.\footnote{A super-resolution image provides more image detail than each image of the original image sequence.}

A survey of more work on image processing in neuroendoscopy is found in \cite{Kon16-Survey}.

\section{Conclusion}
This technical report presented a short introduction into the logarithmic search algorithm for image mosaics. A few iterations of Kourogi's algorithm were used to initialize the motion field, but the central motion field estimation with logarithmic search requires no iterations. In a variety of applications \citep{Bloemen09,Liebig11Diss,Lieb14} it was shown to operate robustly and fast. 

The PhD thesis of 
\citet{Liebig11Diss} and the paper \citet{Lieb14} investigated image mosaics based on Kourogi's algorithm  and on the logarithmic search method in neuroendoscopy under clinical conditions. With large image sizes (720x576 pixels) and color image processing, a frame rate of 3-4 fps could be established.  Three clinical observers graded the quality of the image mosaic in terms of usefulness for surgery. It was found that the logarithmic search method showed significantly better results than Kourogi's method. The logarithmic search method was reliable in operation and provided higher-quality mosaics. There is however a need for faster frame rates to create a smooth panorama image for practical use during surgery.

\end{document}